\documentclass[a4paper]{article}

\usepackage{INTERSPEECH2022}
\usepackage{amsmath}
\usepackage{booktabs}
\usepackage{graphicx}
\usepackage{url}

\IfFileExists{microtype.sty}{%
  \usepackage{microtype}
  \microtypesetup{protrusion=true,expansion=true}
}{}

\newcommand{\inlineheading}[1]{%
  \par\smallskip\noindent\textbf{#1}\enspace\ignorespaces%
}

\title{Leading-Silence Augmentation and Multi-Stage Synthetic Supervision for the Second MLC-SLM Challenge}
\name{Kexin Shi$^{1,2*}$, Renhe Sun$^{1*}$, Yuge Huang$^1$, Ximeng Wang$^2$, Jiayi Zhou$^1$, Jian Liu$^1$, Malu Zhang$^2$}

\address{
  $^1$Ant Group,
  $^2$UESTC
  }
\email{skx534179@antgroup.com, sunrenhe.srh@antgroup.com}

\begin{document}

\maketitle

\begingroup
\renewcommand{\thefootnote}{}
\footnotetext{
$^{*}$These authors contributed equally to this work.
}
\endgroup

\begin{abstract}
The second Multilingual Conversational Speech Language Model (MLC-SLM) Challenge evaluates two tasks over complete, unsegmented multilingual conversations: speaker diarization and recognition (Task~1) and conversational speech understanding (Task~2). Neither task provides oracle utterance boundaries or speaker labels at evaluation, and Task~2 provides no question--answer training set. For Task~1, we fine-tune VibeVoice-ASR-7B with random leading-silence cropping, consistent timestamp correction, and an exponential moving average (EMA) training strategy. For Task~2, we construct synthetic question--answer pairs through multimodal candidate generation, silent-audio filtering, and distribution-matched augmentation, and fine-tune Qwen3-Omni-30B-A3B-Instruct for tagged direct answering. On the Task~1 evaluation set, cropping reduces tcpMER from 18.30\% to 17.27\%, and EMA further reduces it to 16.73\%. On the Task~2 evaluation set, jointly applying distribution-matched augmentation and tagged direct answering raises accuracy from 83.0\% to 86.0\%.
\end{abstract}


\section{Introduction}

\begin{figure*}[!t]
    \centering
    \IfFileExists{image.png}{%
        \includegraphics[width=0.92\textwidth,keepaspectratio]{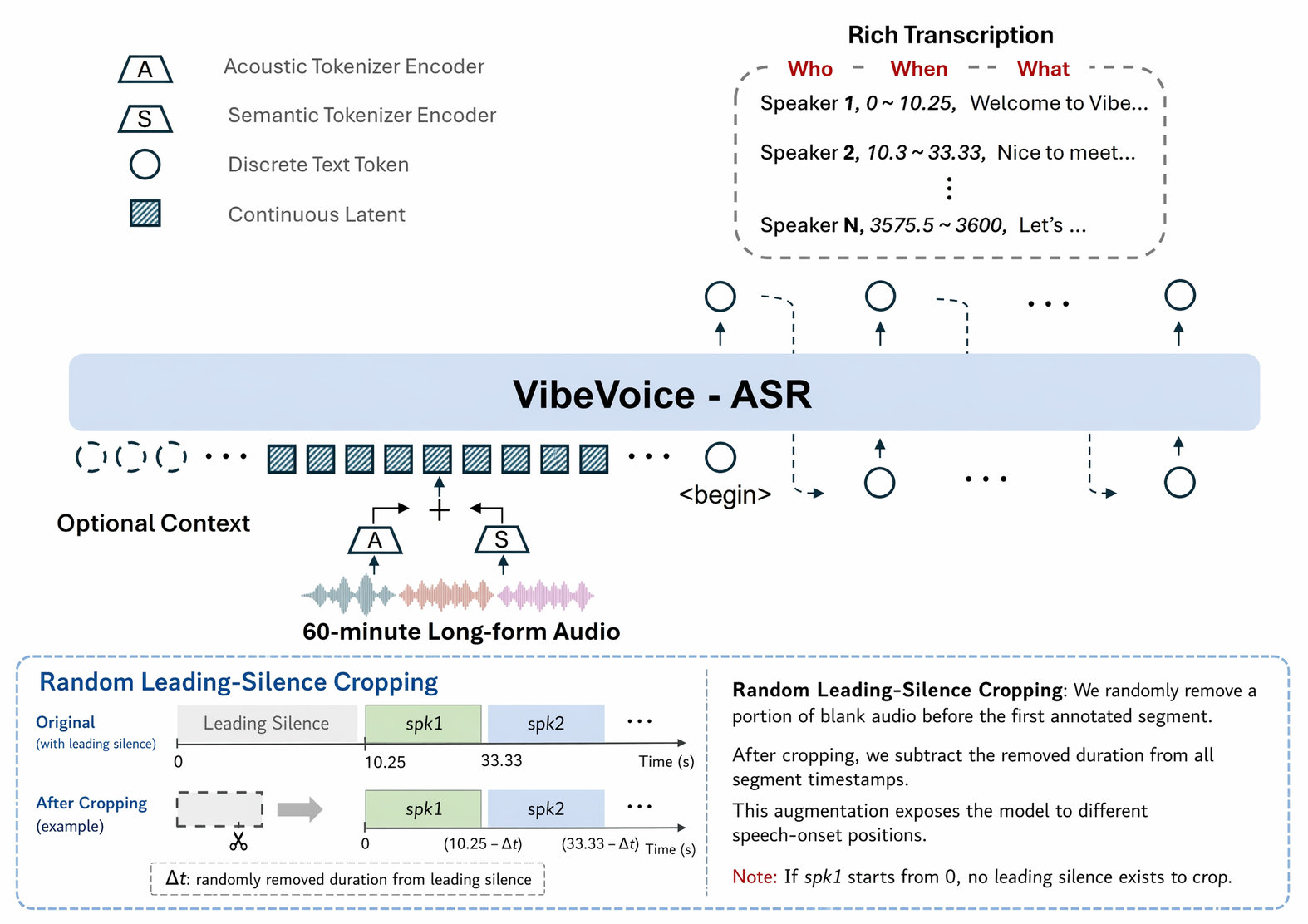}%
    }{%
        \fbox{\parbox[c][0.12\textheight][c]{0.88\textwidth}{\centering Placeholder for \texttt{image.png}: the original VibeVoice-ASR backbone with random leading-silence cropping and consistent timestamp correction.}}%
    }
    \caption{Overview of our Task~1 system, which retains the original VibeVoice-ASR backbone and adds random leading-silence cropping with consistent timestamp correction.}
    \label{fig:task1_system}
\end{figure*}

The second Multilingual Conversational Speech Language Model (MLC-SLM) Challenge evaluates two tasks over complete, unsegmented multilingual conversations. Task~1 combines speaker diarization and recognition, whereas Task~2 evaluates acoustic, semantic, and joint acoustic--semantic understanding through single-answer multiple-choice questions. At evaluation, neither task provides oracle utterance boundaries or speaker labels, and Task~2 provides only development examples rather than a question--answer training set~\cite{mlcslm2026}. These settings combine long-form speaker-attributed transcription with supervision-scarce conversation-level audio understanding.

Task~1 is closely related to end-to-end speaker-attributed ASR, which jointly predicts transcription and speaker information and can be extended to continuous recordings and word-level timing~\cite{kanda2021saasr,kanda2021continuous,kanda2022transcribe}. Long recordings without oracle utterance boundaries remain especially challenging~\cite{mao2020longconversation}. We retain the single-pass VibeVoice-ASR-7B backbone~\cite{peng2026vibevoice} and modify only fine-tuning: random leading-silence cropping varies the temporal origin, consistent timestamp correction preserves annotation alignment, and an EMA training strategy tracks the trainable parameters.

Task~2 must construct task-matched supervision from the released audio. Long-form audio models still struggle with localization, temporal reasoning, counting, and non-phonemic information on full-length recordings~\cite{ahia2025blab}. Large synthetic audio-QA corpora and long-audio QA data have supported instruction-tuned audio understanding~\cite{gong2024ltu,ghosh2025audioflamingo2}, but audio-contribution filtering shows that some multiple-choice examples remain answerable when the audio is replaced by silence~\cite{he2026audiocontribution}. We therefore generate candidate question--answer pairs, apply silent-audio filtering and distribution-matched augmentation, and fine-tune Qwen3-Omni-30B-A3B-Instruct~\cite{xu2025qwen3} for tagged direct answering.

We use one submitted prediction model per task, without model-, score-, or output-level fusion. Our main contributions and findings are:
\begin{itemize}
    \item In the cumulative Task~1 ablation, the LoRA baseline obtains 18.30\% tcpMER; adding leading-silence cropping with timestamp correction yields 17.27\%, and adding EMA on top yields 16.73\%.
    \item For Task~2, the pipeline produces approximately 127k synthetic examples. On evaluation set, the final cumulative configuration with distribution-matched augmentation and tagged direct answering raises accuracy from 83.0\% to 86.0\%.
\end{itemize}

\section{System Description}

\subsection{Task 1: Multilingual Conversational Speech Diarization and Recognition}

\subsubsection{Backbone and Task Formulation}

We use VibeVoice-ASR-7B as the Task~1 backbone. The model supports recordings of up to 60 minutes, more than 50 languages, and code-switching~\cite{peng2026vibevoice}, making it suitable for multilingual conversational speech. We fine-tune it on complete challenge recordings and serialize its predictions directly in the official output format.

We retain the original architecture and single-pass formulation. In a single pass, the model encodes each complete recording and emits a temporally ordered sequence of transcripts, speaker identities, and timestamps. Our changes affect only fine-tuning and leave the decoder architecture and output format unchanged. Figure~\ref{fig:task1_system} summarizes the system.

\subsubsection{Random Leading-Silence Cropping}

Established ASR augmentation methods perturb waveform speed~\cite{ko2015audioaugmentation} or mask time--frequency regions in acoustic features~\cite{park2019specaugment}. Random leading-silence cropping instead targets variation in the temporal origin and leading non-speech context while preserving the annotated speech. Because the amount of non-speech audio before the first annotated utterance varies across recordings, we randomly remove part of this region during fine-tuning. For a recording of duration $T$ whose first annotated segment begins at $s_1$, we sample a crop duration $\delta$ according to the training-time cropping policy, subject to $0 \leq \delta \leq s_1$, and remove the interval $[0,\delta)$.

After cropping, we shift every annotated timestamp by the same offset:
\begin{equation}
    s_i' = s_i - \delta, \qquad
    e_i' = e_i - \delta, \qquad
    T' = T - \delta.
\end{equation}
The constraint $\delta \leq s_1$ ensures that no annotated speech is removed. A common shift preserves transcripts, speaker labels, turn order, segment durations, and relative timing; only the leading non-speech context, absolute time origin, and total recording duration change.

This augmentation changes the temporal presentation of an existing example without synthesizing a new conversation or perturbing internal speaker-turn boundaries. Consequently, the aggregate ablation measures the effect of leading-silence augmentation but does not directly establish robustness to turn-boundary errors.

\begin{figure*}[!t]
    \centering
    \IfFileExists{figs/mlc-slm-data.pdf}{%
        \includegraphics[width=\textwidth,keepaspectratio]{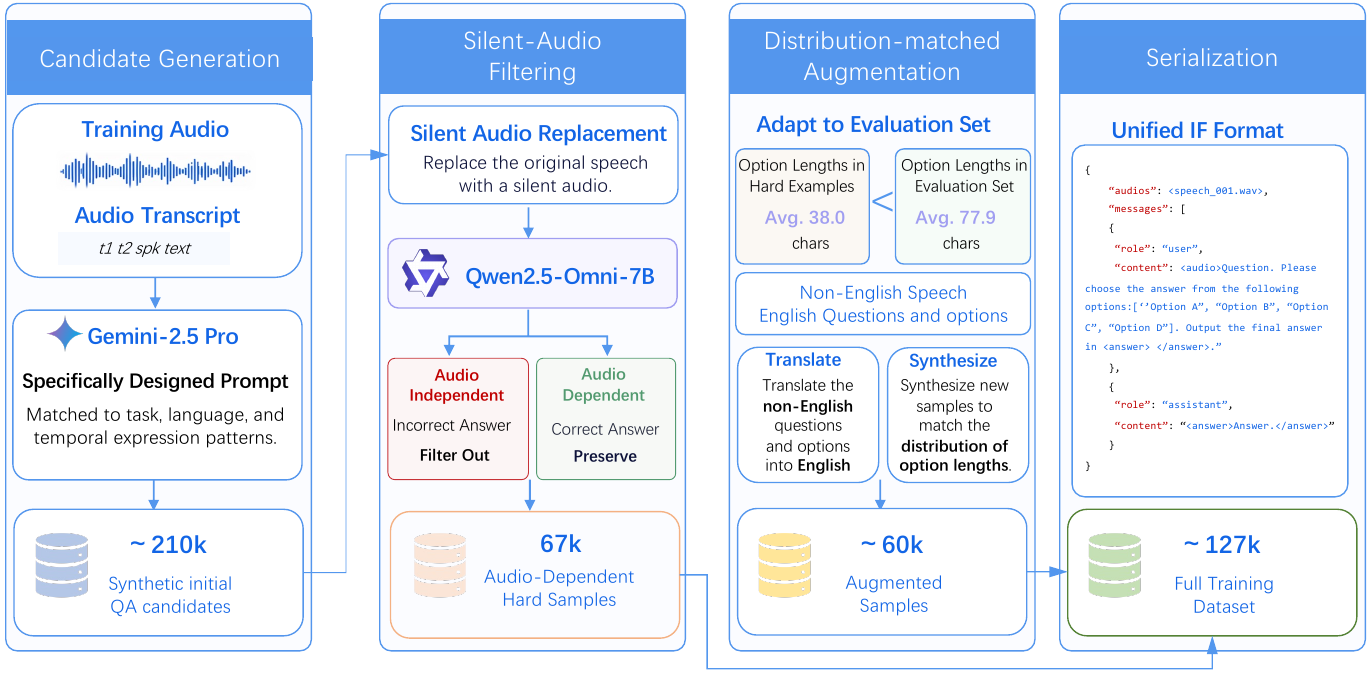}%
    }{%
        \fbox{\parbox[c][0.13\textheight][c]{0.92\textwidth}{\centering Placeholder for \texttt{figs/mlc-slm.pdf}: candidate QA generation, silent-audio filtering, distribution-matched augmentation, and instruction-format serialization.}}%
    }
    \caption{Task~2 synthetic-supervision pipeline: candidate generation, silent-audio filtering, distribution-matched augmentation, and unified instruction-format serialization.}
    \label{fig:data_pipeline}
\end{figure*}

\subsubsection{EMA Training Strategy}

During fine-tuning, we maintain an exponential moving average of the trainable parameters~\cite{moralesbrotons2024ema}. After optimizer step $t$, the EMA state is updated as
\begin{equation}
    \bar{\boldsymbol{\theta}}_{t}
    = \lambda \bar{\boldsymbol{\theta}}_{t-1}
    + (1-\lambda)\boldsymbol{\theta}_{t},
\end{equation}
where $\boldsymbol{\theta}_{t}$ denotes the current trainable parameters, $\bar{\boldsymbol{\theta}}_{t}$ denotes the EMA state, and $\lambda$ is the decay coefficient. We set $\lambda=0.99$ and update the EMA state after every optimizer step throughout fine-tuning.

EMA is applied throughout fine-tuning and does not alter decoding or output serialization. Table~\ref{tab:task1_results} reports the cumulative configuration in which EMA is added after random leading-silence cropping.

\subsubsection{Training Configuration}

We fine-tune VibeVoice-ASR-7B using low-rank adaptation (LoRA)~\cite{hu2022lora}. The LoRA rank, scaling factor, and dropout probability are set to 32, 128, and 0.05, respectively, and training runs for five epochs. Each of the 16 data-parallel workers uses a per-device batch size of 1 with four gradient-accumulation steps, resulting in an effective global batch size of 64.

The optimization configuration uses a learning rate of $1\times10^{-4}$, a warmup ratio of 0.03, weight decay of 0.01, and gradient clipping with a maximum norm of 1.0. We reduce memory consumption through gradient checkpointing~\cite{chen2016checkpointing}, bfloat16 arithmetic, and DeepSpeed ZeRO-2~\cite{rajbhandari2020zero}. Random leading-silence cropping, timestamp correction, and the EMA training strategy remain enabled throughout fine-tuning.

\subsection{Task 2: Multilingual Conversational Speech Understanding}

\subsubsection{Foundation Model and Prediction Format}

Qwen2-Audio demonstrates general audio analysis and instruction-conditioned interaction within the Qwen audio-language model line~\cite{chu2024qwen2audio}. Following this broader trend, our Task~2 system uses Qwen3-Omni-30B-A3B-Instruct as its sole prediction model~\cite{xu2025qwen3}. Each instance is formatted as a user message containing the complete conversation audio, one single-answer question, two to four candidate options, and an instruction to enclose the selected answer within \texttt{<answer></answer>} tags.

The model processes the original conversation audio directly, without relying on an externally generated transcript. It therefore retains access to both lexical content and non-lexical acoustic cues. The tagged answer format provides a consistent supervision target and facilitates uniform prediction extraction during evaluation. Figure~\ref{fig:data_pipeline} summarizes the complete Task~2 pipeline.

\subsubsection{Synthetic Training Data}

\inlineheading{Candidate generation.}
Synthetic instruction generation offers a scalable alternative when task-specific supervision is scarce~\cite{wang2023selfinstruct}, and prior audio-language work has extended this strategy to large question--answer corpora~\cite{gong2024ltu}. We use Gemini 2.5 Pro~\cite{comanici2025gemini} to generate approximately 210k candidate question--answer pairs from the released training audio. The prompts follow the released development examples in question type, language, timestamp style, and the required two-to-four-option, single-answer format.

\inlineheading{Silent-audio filtering.}
Some generated questions may be answerable from textual priors or option artifacts rather than evidence in the associated conversation. Audio-language models can favor textual input when audio and text disagree~\cite{wang2025textbias}, and recent audio-QA work has identified multiple-choice examples with little or no audio contribution and used audio-contribution filtering to separate them~\cite{he2026audiocontribution}. To reduce such cases, we use Qwen2.5-Omni-7B~\cite{xu2025qwen25omni} as a counterfactual filter. For each candidate, we retain the question, options, and expected answer but replace the original audio with silence. We discard candidates that are still answered correctly and retain those for which the prediction becomes incorrect. This procedure retains approximately 67k pairs, or 32\% of the candidate pool. It increases the likelihood that retained questions depend on audio evidence but does not by itself guarantee acoustic grounding.

\inlineheading{Distribution-matched augmentation.}
Translation-based multilingual AQA has previously expanded an audio-QA corpus across eight question languages~\cite{behera2023multilingualaqa}. Relative to the released development examples, the evaluation inputs contain longer options on average (77.9 versus 38.0 characters) and pair some non-English conversations with English questions and options. Using these aggregate, label-free input properties, we translate selected training questions and synthesize long-option examples, producing approximately 60k additional instances. No question, option set, or answer from the evaluation set is copied into the training data.

\inlineheading{Serialization.}
All retained and augmented examples are converted to a common chat-style schema. During training, the user message contains the audio placeholder, question, candidate options, and output instruction, whereas the assistant message contains only the gold answer enclosed by \texttt{<answer></answer>} tags. At evaluation, the model receives only the user message.

\subsubsection{Training, Inference, and Challenge Compliance}

We fine-tune Qwen3-Omni using LoRA~\cite{hu2022lora}. The LoRA parameters, visual transformer (ViT), and modality aligner remain trainable, while all other parameters are frozen. Training uses a learning rate of $5\times10^{-5}$ for two epochs, with a micro-batch size of 4 and a global batch size of 64 across 16 accelerator devices.

At inference, sampling is disabled and the implementation sets the temperature to 0.0. The text enclosed by \texttt{<answer></answer>} tags is extracted as the prediction.

Each task submission is produced by a single trained model, without model-, score-, or output-level fusion. Gemini 2.5 Pro and Qwen2.5-Omni are used exclusively in the offline Task~2 data-construction pipeline, whereas every submitted Task~2 prediction is generated solely by Qwen3-Omni. The Qwen checkpoints are publicly released~\cite{xu2025qwen3,xu2025qwen25omni} and disclosed; Gemini 2.5 Pro is used only under the permitted Task~2 commercial-API exception~\cite{mlcslm2026}.

\section{Experiments and Results}

\subsection{Evaluation Protocol}

For Task~1, we use the official scorer and baseline configuration~\cite{mlcslm2026task1baseline}. Diarization error rate (DER) first determines the optimal permutation between reference and hypothesized speakers. After speaker mapping, the recognition streams are evaluated using tcpCER for Japanese, Korean, and Thai and tcpWER for all other languages with the official 5-s collar~\cite{mlcslm2026task1baseline}; metric computation uses MeetEval~\cite{vonneumann2023meeteval,vonneumann2025werdefinitions}. We report the across-language average as tcpMER, where lower indicates better performance.

Task~2 evaluates conversation understanding using single-answer multiple-choice accuracy. Each question contains two to four options with exactly one correct answer and may require acoustic, semantic, or joint acoustic--semantic evidence. Only the evaluation score contributes to the final challenge ranking~\cite{mlcslm2026}.

\subsection{Task 1 Results}

Table~\ref{tab:task1_results} summarizes three cumulative Task~1 configurations evaluated with the official tcpMER metric, for which lower values are better.

\begin{table}[t]
\centering
\caption{Cumulative Task~1 ablation under the official tcpMER protocol. Lower is better.}
\label{tab:task1_results}
\setlength{\tabcolsep}{5pt}
\renewcommand{\arraystretch}{1.08}
\begin{tabular}{@{}lr@{}}
\toprule
System configuration & tcpMER (\%) $\downarrow$ \\
\midrule
LoRA baseline (ours) & 18.30 \\
$+$ random leading-silence cropping & 17.27 \\
$+$ EMA training strategy & \textbf{16.73} \\
\bottomrule
\end{tabular}
\end{table}

The configurations and their observed changes are as follows:
\begin{itemize}
    \item \textbf{LoRA baseline (ours):} We fine-tune VibeVoice-ASR-7B using LoRA while retaining the original architecture, single-pass decoding procedure, and serialized output format. This reference configuration obtains a tcpMER of 18.30\%.
    \item \textbf{+ Random leading-silence cropping:} We randomly crop part of the non-speech region preceding the first annotated segment; the required timestamp correction shifts every annotation by the same offset to preserve alignment. This configuration obtains 17.27\% tcpMER, 1.03 absolute tcpMER points lower than the LoRA baseline, corresponding to a 5.6\% relative reduction.
    \item \textbf{+ EMA training strategy:} During fine-tuning, we additionally maintain an exponential moving average of the trainable parameters and update it after every optimizer step; this strategy does not modify decoding. With cropping already enabled, the EMA configuration obtains 16.73\% tcpMER, a further reduction of 0.54 absolute tcpMER points, or 3.1\% relative to the preceding configuration.
\end{itemize}

The two successive additions are associated with monotonically lower tcpMER in the evaluated sequence, and cropping gives the larger observed incremental reduction. Overall, the complete configuration records a 1.57-point absolute decrease from the LoRA baseline, equivalent to an 8.6\% relative reduction, while retaining the original single-pass inference procedure.

Because the ablation rows are cumulative, these comparisons remain conditional. The cropping row evaluates cropping together with the timestamp correction required to keep its labels valid. The EMA difference is measured only after cropping is enabled; without an EMA-only row, the EMA effect and any interaction between the two training strategies cannot be estimated. In addition, aggregate tcpMER cannot show whether the reductions are uniform across languages, recording durations, or first-speech onset conditions; per-language and onset-conditioned analyses would be required to localize the gains.

\subsection{Task 2 Results}

\begin{table}[t]
\centering
\caption{Task~2 ablation of distribution-matched augmentation and response format on evaluation set (accuracy, \%).}
\label{tab:task2_ablation}
\begin{tabular}{lc}
\toprule
Configuration & Accuracy (\%) $\uparrow$ \\
\midrule
Baseline (direct answering) & 78.0 \\
+ candidate generation & 81.0 \\
+ silent-audio filtering & 83.0 \\
+ distribution-matched augmentation & 85.0 \\
+ tagged direct answering & \textbf{86.0} \\
\bottomrule
\end{tabular}
\end{table}

Table~\ref{tab:task2_ablation} presents the incremental ablation results for Task~2 on evaluation set. Starting from the direct-answering baseline, we progressively add candidate generation, silent-audio filtering, distribution-matched augmentation, and the tagged direct-answering format. Accuracy improves from 78.0\% to 86.0\%, confirming that the proposed synthetic data construction pipeline and response-format design provide complementary gains.

\inlineheading{Effect of candidate generation and filtering.}
Candidate generation improves accuracy from 78.0\% to 81.0\%, indicating that the synthetic QA pairs provide useful task-specific supervision. Applying silent-audio filtering on top of the generated candidates further raises accuracy from 81.0\% to 83.0\%. This result suggests that filtering out samples that remain answerable with silent audio helps retain more audio-dependent examples and improves the effectiveness of the training data.

\inlineheading{\mbox{Effect of distribution-matched augmentation.}} 
Additionally, distribution-matched augmentation on top of the filtered data further improves accuracy from 83.0\% to 85.0\%. This gain shows that better matching the training data to the evaluation data distribution leads to stronger generalization.

\inlineheading{Effect of response format.}
Finally, introducing the tagged direct-answering format improves accuracy from 85.0\% to 86.0\%. Besides the performance gain, the explicit answer tags also make answer extraction more reliable during inference. We therefore use this format in the final submitted system.





\section{Conclusion}

We presented independently adapted single-model systems for both tasks of the second MLC-SLM Challenge. For Task~1, random leading-silence cropping, consistent timestamp correction, and an EMA training strategy reduce tcpMER from 18.30\% to 16.73\%. For Task~2, we constructed approximately 127k synthetic examples through multimodal candidate generation, silent-audio filtering, and distribution-matched augmentation, and fine-tuned Qwen3-Omni for tagged direct answering. The final sequential configuration reached 86.0\% accuracy, compared with 83.0\% for the reasoning-then-answer baseline.

For Task~1, future work will explore richer speaker--temporal modeling and better use of conversational context. For Task~2, we will investigate stronger multilingual adaptation and more reliable audio-grounded long-context understanding.

\end{document}